\documentclass[sigconf, nonacm]{acmart}
\AtBeginDocument{%
  }

\usepackage{graphicx}
\usepackage{amsmath}
\newcommand{\modelname}{\textsc{MiSo}}
\usepackage{subcaption}
\usepackage{multirow}

\begin{document}
\title{Fine-Scale Soil Mapping in Alaska with Multimodal Machine Learning}

% \author{Yijun Lin\textsuperscript{*1}, Theresa Chen\textsuperscript{*1}}
% \affiliation{
%   \institution{\textsuperscript{1}University of Minnesota}
%   \country{Minneapolis-St. Paul, MN, USA}}

% --- Group authors by affiliation ---
\author{Yijun Lin\textsuperscript{*}, Theresa Chen\textsuperscript{*}}
\affiliation{%
  \institution{University of Minnesota}
  \city{Minneapolis-St. Paul}
  \state{MN}
  \country{USA}
}
\email{{*lin00786, *chen7924}@umn.edu}

\author{Colby Brungard}
\affiliation{%
  \institution{New Mexico State University}
  \city{Las Cruces}
  \state{NM}
  \country{USA}
}
\email{cbrung@nmsu.edu}

\author{Grunwald Sabine}
\affiliation{%
  \institution{University of Florida}
  \city{Gainesville}
  \state{FL}
  \country{USA}
}
\email{sabgru@ufl.edu}

\author{Sue Ives, Matt Macander}
\affiliation{%
  \institution{ABR, Inc.}
  \city{Anchorage and Fairbanks}
  \state{AK}
  \country{USA}
}
\email{{sives,mmacander}@abrinc.com}

\author{Timm Nawrocki}
\affiliation{%
  \institution{University of Alaska-Anchorage}
  \state{AK}
  \country{USA}
}
\email{twnawrocki@alaska.edu}

\author{Yao-Yi Chiang, Nic Jelinski}
\affiliation{%
  \institution{University of Minnesota}
  \city{Minneapolis-St. Paul}
  \state{MN}
  \country{USA}
}
\email{{yaoyi, jeli0026}@umn.edu}

%   \institution{\textsuperscript{2}University of Minnesota, Minneapolis-St. Paul, MN, USA}
%   \institution{\textsuperscript{2}New Mexico State University, Las Cruces, NM, USA}
%   \institution{\textsuperscript{3}University of Florida, Gainesville, FL, USA}
%   \institution{\textsuperscript{4}ABR, Inc., Anchorage and Fairbanks, AK, USA}
%   \institution{\textsuperscript{5}University of Alaska-Anchorage, AK, USA}

\renewcommand{\shortauthors}{Yijun Lin\textsuperscript{*}, Theresa Chen\textsuperscript{*}, and et al.}

\thanks{\textsuperscript{*}Both authors contributed equally to this research.}

\begin{abstract}
Fine-scale soil mapping in Alaska, traditionally relying on fieldwork and localized simulations, remains a critical yet underdeveloped task, despite the region’s ecological importance and extensive permafrost coverage. As permafrost thaw accelerates due to climate change, it threatens infrastructure stability and key ecosystem services, such as soil carbon storage. High-resolution soil maps are essential for characterizing permafrost distribution, identifying vulnerable areas, and informing adaptation strategies. We present \modelname, a vision-based machine learning (ML) model to produce statewide fine-scale soil maps for near-surface permafrost and soil taxonomy. The model integrates a geospatial foundation model for visual feature extraction, implicit neural representations for continuous spatial prediction, and contrastive learning for multimodal alignment and geo-location awareness. We compare \modelname~with Random Forest (RF), a traditional ML model that has been widely used in soil mapping applications. Spatial cross-validation and regional analysis across Permafrost Zones and Major Land Resource Areas (MLRAs) show that \modelname~generalizes better to remote, unseen locations and achieves higher recall than RF, which is critical for monitoring permafrost thaw and related environmental processes. These findings demonstrate the potential of advanced ML approaches for fine-scale soil mapping and provide practical guidance for future soil sampling and infrastructure planning in permafrost-affected landscapes. The project will be released at \url{https://github.com/knowledge-computing/Peatland-permafrost}.

\end{abstract}

\maketitle

\section{Introduction}

% Accurate fine-scale soil prediction, or digital soil mapping, is essential for a wide range of applications, including environmental planning, natural resource management, and infrastructure development~\cite{lal2017soil,lal2021soils}. While many soil mapping products have been developed at national~\cite{reybold1989soil,bliss1995preparing,sanchez2009digital,arrouays2014globalsoilmap} and local~\cite{shi2004case} scales for the continental United States, few efforts have focused on Alaska due to its vast, remote, and sparsely populated landscape. However, Alaska is ecologically important, with heterogeneous soil types and extensive permafrost coverage, making accurate soil mapping particularly critical.

Soil maps are essential resources for a wide range of applications, such as environmental planning, natural resource management, and infrastructure development~\cite{lal2017soil,lal2021soils}. While many soil mapping products have been developed at national~\cite{reybold1989soil,bliss1995preparing,sanchez2009digital,arrouays2014globalsoilmap} and local~\cite{shi2004case} scales for the continental United States, few efforts have focused on Alaska due to its vast, remote, and sparsely populated landscape. However, soil maps are critically important to Alaska, serving as essential tools for modeling heterogeneous soils and understanding the extent of permafrost coverage.

Near-surface permafrost (NSP)---soil that remains at or below 0\,${}^{\circ}$C for at least two consecutive years within 1-2 meters of the soil surface---is important for maintaining landscape stability and carbon storage~\cite{jorgenson2013reorganization,balshi2009vulnerability}. With rising global temperatures, permafrost is undergoing rapid thaw and degradation, posing significant risks to homes, roads, and critical infrastructure in Alaska~\cite{chapin2000arctic,hjort2022impacts,jorgenson2006abrupt,shur2007patterns}. Existing statewide soil mapping products, such as STATSGO2~\cite{soil2022Statsgo2} and gNATSGO~\cite{soil2022gridded}, lack the spatial resolution and details for representing micro-topographic and localized environmental assessments. Therefore, developing high-resolution spatial data on NSP and related soil taxonomy (e.g., the Gelisols order, which is highly correlated with permafrost) is essential for accurately modeling soil dynamics and supporting targeted environmental planning~\cite{jelinski2024estimates}.

Detecting and monitoring NSP and soil taxonomy remains challenging due to their subsurface nature, spatial heterogeneity, and prevalence in remote and inaccessible regions~\cite{riseborough2008recent}. Several studies have applied process-based or transient models to simulate permafrost dynamics at high spatial resolutions for small areas ($<$25,000km$^2$)~\cite{panda2014higha,panda2014highb}. However, these models often rely on environmental covariates and physical assumptions that may not hold uniformly in space or are only available in limited locations across extremely large areas like Alaska~\cite{lawrence2005projection, lawrence2008sensitivity, lawrence2012simulation}.

% Statistical and machine learning models have been increasingly applied to statewide soil mapping by capturing complex, nonlinear interactions between soil characteristics and ecological or environmental predictors~\cite{panda2012near,panda2010remote,pastick2015distribution}. One of the most popular approaches is Random Forest which has been used in both the production of widely used large-scale 100-m soil maps over the entire United States~\cite{ramcharan2018soil, nauman2024soil} as well as finer, 30-m soil property maps, in Alaska and elsewhere \cite{pastick2015distribution, thaler2023high, nitze2018remote}. Recent advances in deep learning and computer vision have shown strong potential in soil modeling, though most existing work focus on broader tasks using convolutional~\cite{kakhani2024ssl,wadoux2019multi,padarian2019using}, with limited emphasis on fine-scale soil prediction. Moreover, the computer vision community has developed geo-foundation models~\cite{bastani2023satlaspretrain,cong2022satmae,hong2024spectralgpt}, typically pretrained on large-scale satellite imagery, which have achieved state-of-the-art performance in various geospatial tasks~\cite{lambhate2024finetuning,xiao2024foundation,bastani2023satlaspretrain}. However, these models remain underutilized in soil science, potentially due to computational constraints, challenges in task-specific adaptation, and limited interdisciplinary collaboration.

Statistical and machine learning (ML) models have been increasingly applied to soil mapping by leveraging ecological or environmental predictors, such as remote sensing data, to capture complex, nonlinear relationships with soil properties~\cite{panda2012near,panda2010remote,pastick2015distribution}. One of the most commonly used methods is Random Forest, adapting in both national-scale 100-meter soil maps over the conterminous United States~\cite{ramcharan2018soil,nauman2024soil} and fine-resolution, 30-meter soil maps in Alaska and elsewhere~\cite{pastick2015distribution, thaler2023high, nitze2018remote}. Recent advances in deep learning and computer vision have demonstrated strong capability in extracting spatial features from imagery and have been applied to a range of broader soil-related tasks, e.g., soil carbon prediction~\cite{kakhani2024ssl,padarian2019using}. Moreover, the computer vision community has developed geo-foundation models~\cite{bastani2023satlaspretrain,cong2022satmae,hong2024spectralgpt}, typically pretrained on large-scale satellite imagery, which have achieved state-of-the-art performance in various geospatial tasks~\cite{xiao2024foundation,bastani2023satlaspretrain}. However, these models remain underutilized in soil science, potentially due to challenges in task-specific adaptation and limited interdisciplinary collaboration.

This paper investigates the utilization of multimodal datasets and advanced computer vision techniques to generate fine-scale soil maps for (1) near-surface permafrost, and (2) soil taxonomy in Alaska. We introduce \modelname\footnote{\modelname~stands for \textbf{M}ult\textbf{I}modal ML for Fine-Scale \textbf{SO}il Mapping}~that integrates a pretrained geospatial foundation model based on the SWIN Transformer~\cite{bastani2023satlaspretrain}, implicit image functions for continuous spatial prediction~\cite{chen2021learning}, and contrastive learning for multimodal feature alignment and geo-location awareness. The proposed architecture is highly effective for handling sparse observations, multimodal raster inputs, and making soil predictions at arbitrary locations.
For comparison, we utilize Random Forest (RF), due to its frequent use in the soil science community. While RF could perform well under standard evaluation metrics, its predictions tend to be biased toward non-NSP or dominant soil classes that are typically of less interest to soil scientists. In contrast, \modelname~produces more conservative NSP probability estimates, which are particularly valuable for guiding field sampling and early-warning assessments, and also achieve better performance in predicting the spatial distribution of detailed soil taxonomy than RF. 
% While RF could produce accurate results and capture broad spatial patterns for NSP and soil taxonomy, it tended to be biased towards dominant soil classes,  limiting its ability to represent heterogeneous soil profiles, which are important to representing local microclimates in numerous fine-scale soil modeling tasks \cite{thompson2012harmonization, yi2018characterizing}. In contrast, \modelname~produces more conservative probability estimates for NSP presence and better performance in understanding the distribution of detailed soil taxonomy than RF. 
Through a comprehensive comparison of \modelname~and RF, we provide guidance on the strengths and limitations of each method and suggest appropriate use cases depending on the scientific modeling goals, application needs, and data constraints. 

\vspace{0.05in}
The main contributions of our paper include:
\begin{itemize}
    \item Introducing \modelname, a model that integrates state-of-the-art geo foundation models with implicit image functions for fine-scale soil mapping, enabling direct predictions from sparse field observations without the need to predefine a fixed output resolution.
    \item Producing the first 10 meter/pixel soil maps for near-surface permafrost presence-absence and soil taxonomy classification across Alaska with ~\modelname~and an RF model. 
    \item Providing a comprehensive comparison and analysis between ~\modelname~and RF to help end users understand the conditions under which each model may be preferred. 
\end{itemize}

% Methods such as Random Forests have demonstrated promise by learning complex relationships between soil properties and widely available environmental features, such as satellite-derived indices, terrain metrics, and climate variables. These models have improved prediction accuracy and scalability in data-rich environments. However, they still suffer from limitations. Chief among them is the reliance on manual feature engineering, where expert knowledge is required to select and aggregate relevant variables. In addition, many machine learning models require the definition of spatial buffers to represent a location's context, which can introduce biases or inconsistencies across regions with varying data density and landscape heterogeneity. This novel study expands upon decades of permafrost-related research by producing the first medium-resolution (30 m × 30 m) maps of near-surface (within 1 m) permafrost (NSP) and associated uncertainty estimates, throughout all of mainland Alaska. 

\section{Related Work}
\subsection{Soil Mapping Products in Alaska}
The Alaska State Soil Geographic Database (STATSGO2)\cite{soil2022Statsgo2}, developed by the USDA Natural Resources Conservation Service (NRCS), is currently the only statewide soil product available for Alaska. STATSGO2 provides a statewide soil resource overview at a spatial scale of 1:1,000,000, where individual map units can span tens to hundreds of square kilometers and combine diverse soil types. The gridded National Soil Survey Geographic Database (gNATSGO)~\cite{soil2022gridded} offers a 10-meter resolution soil map by integrating STATSGO2 with the Soil Survey Geographic Database (SSURGO). However, due to limited SSURGO coverage in Alaska, gNATSGO is largely derived from the coarse resolution of STATSGO2, and the 10-meter resolution reflects rasterization rather than true spatial details \cite{soil2022gridded}. These limitations highlight the need for new high-resolution soil mapping products for Alaska. 

\subsection{Process-Based Models}
Process-based models rely on a combination of governing equations that specify interactions between physical, chemical, and biological variables as well as domain knowledge on how certain soil characteristics respond to environmental conditions (e.g., \cite{jafarov2012numerical, marchenko2001model, lawrence2019community}). For instance, the Geophysical Institute Permafrost Lab (GIPL) models for simulating soil temperature dynamics and, subsequently, permafrost thaw work by solving the heat equation using environmental variables that affect soil temperature, such as air temperature and precipitation, snow cover, and surface geology \cite{jafarov2012numerical}. Other standard process-based models include several types of global climate models, such as Community Land Models and Community Climate System Models~\cite{blackmon2001community}, which can be used to tie interactions from the atmosphere, ocean, and earth to permafrost freeze/thaw cycles \cite{lawrence2005projection, lawrence2008sensitivity, lawrence2012simulation}. Variants of process-based models are generally used for coarse-scale (>1km $\times$ 1km) permafrost modeling~\cite{jafarov2012numerical, marchenko2008numerical}. Process-based models have also been applied to fine-scale modeling tasks in small areas ($<$25,000 km$^2$)~\cite{zhang2012modelling, panda2014high}, but these models are not scalable to large areas as they require extensive amounts of detailed data on soil properties that must be obtained through field work. Generally, process-based models require detailed climate and pedological information as preconditions and, therefore, cannot be applied broadly over a large geographic region at a high resolution (<100 meters).

\subsection{Machine Learning Models}
To mitigate the limitations in process-based models, several studies have leveraged widely available remote sensing data in combination with machine learning models, such as piecewise regression \cite{gangodagamage2014extrapolating}, support vector machines \cite{deluigi2017data}, neural networks \cite{liu2022interannual}, and random forests \cite{pastick2015distribution}, to predict various soil properties. For example, \citet{pastick2015distribution} develop a statewide map of NSP presence over Alaska at 30-m resolution by integrating topographic and climate covariates with statewide field observations using a Random Forest model. \citet{thaler2023high} compare extremely randomized trees, support vector machines, and neural networks in predicting soil characteristics for three small sites on the Seward Peninsula in Alaska, but caution that their models may not generalize well to areas with different ecological and topographical factors than the Seward Peninsula.

Recent advances in deep learning and computer vision have shown promise for modeling complex soil properties. In the Arctic and permafrost research communities, convolutional neural networks (CNNs) have been used to predict permafrost extent and surface soil properties~\cite{langford2019arctic, chen2024retrieving, liu2022interannual}. The Vision Transformer architecture has been used to combine multimodal input data for the soil organic carbon task in non-Arctic regions~\cite{kakhani2024ssl}. In other soil-related domains, such as soil moisture modeling, researchers have integrated computer vision techniques with known governing equations on processes such as water exchange to develop physics-informed machine learning models~\cite{khandelwal2024deepsoil}. However, these approaches are typically tailored to specific soil characteristics and cannot be generalized to broader soil property prediction tasks where the governing equations are unknown. In contrast, this paper introduces a deep learning method that adapts advanced vision transformers to multimodal raster data, capturing rich visual features and their local, spatially variant interactions with terrain information for fine-scale soil mapping.

\subsection{Geospatial Foundation Models}
Geospatial foundation models (GFMs) or vision-based GFMs, typically pretrained on large-scale remote sensing imagery, aim to generate generalizable representations for a variety of downstream geospatial prediction tasks~\cite{cong2022satmae,bastani2023satlaspretrain,jakubik2023foundation,guo2024skysense,nedungadi2024mmearth,xiao2024foundation}. For example, SatMAE~\cite{cong2022satmae} employs a masked autoencoder architecture with a Vision Transformer backbone and is pretrained on multi-spectral satellite images. \citet{ayush2021geography} introduce a contrastive learning approach that aligns satellite images from the same geographic location across time and simultaneously learns to predict image coordinates. In addition to innovations in model architecture, curating diverse and large-scale datasets is also critical to GFMs. SATLAS~\cite{bastani2023satlaspretrain} is a dataset comprising multi-temporal, multi-resolution satellite imagery covering various global regions. The accompanying model, SATLASNET, utilizes a SWIN Transformer~\cite{liu2021swin} to extract image features and is pretrained across multiple geospatial tasks, ranging from image segmentation to classification~\cite{bastani2023satlaspretrain}. Although GFMs have not been widely adopted in soil science, their training strategy and capacity to learn generalizable representations make them well-suited for soil mapping, especially in regions with limited labeled data. This paper proposes incorporating GFMs to extract visual features from satellite imagery and serve as guidance to learn visual representations for other modalities.

\section{Datasets}\label{sec:Datasets}
This section introduces the datasets used as covariates for detailed soil mapping in Alaska. Both field observations and covariates are compiled by the Alaska Soil Data Bank (AKSDB) project.\footnote{\url{https://github.com/alaska-soil-data-bank}}

\subsection{Study Area and Field Observations}
\begin{figure}[h]
    \centering
    \includegraphics[width=\linewidth]{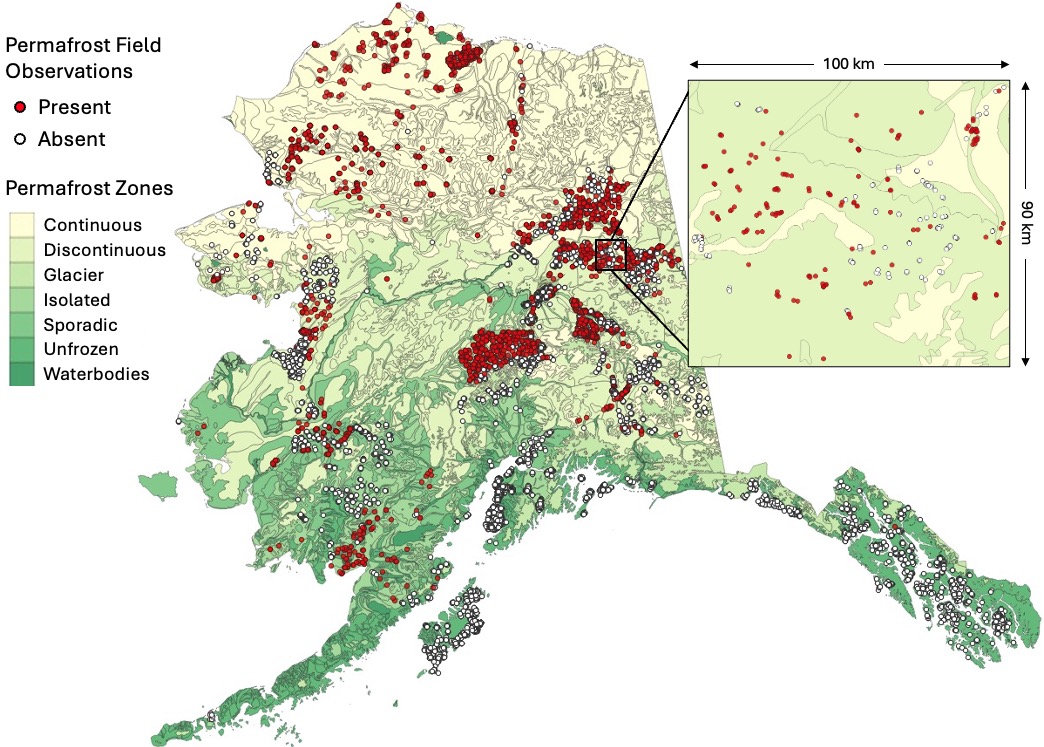}
    \caption{Field observations of NSP presence-absence across Alaska overlaying the map of Permafrost Zones}
    \label{fig:permafrost}
\end{figure}

Alaska (\textasciitilde1,500,000 km$^2$) is the largest and one of the most ecologically diverse states in the U.S. In this work, we utilize a dataset of approximately 38,000 field measurements spanning the entire state of Alaska, sourced from AKSDB, the largest collection of field surveys in Alaska to date. These measurements were compiled from existing but disparate datasets from academia, government agencies and partners, as well as published literature. Due to the diversity of sources, each survey point in the dataset is not guaranteed to have a measurement for every soil property. There are around 15,000 measurements for near-surface permafrost (NSP) and 32,000 for soil taxonomy. Besides soil properties, each survey point has a record of its geographic location (longitude, latitude) and a fieldwork date, which can range from 1952 to 2023 and occur in every month of the year, with the majority occurring after 1998 and during the mid-summer months. While the broad temporal range may cause misalignment between the observations and current environmental conditions, in practice, soil science research typically includes historical observations to enhance spatial coverage for training, particularly in remote and data-sparse regions \cite{ramcharan2018soil, thompson2012harmonization}. Figure~\ref{fig:permafrost} shows the point distribution of NSP observations, which covers all major permafrost zones. These field observations are sparse and unevenly distributed across Alaska. %While the broad temporal range may raise potential concerns about alignment with current environmental conditions, we ultimately include historical observations to enhance spatial coverage for training, especially for remote and data-sparse regions. 

\subsection{Sentinel-2}
Satellite imagery is a valuable source for modeling soil properties, as it captures surface characteristics that indirectly reflect NSP presence or certain soil types, such as vegetation, snow cover, and hydrology. In this paper, we utilize Sentinel-2 multi-spectral imagery provided by AKSDB, covering the entirety of Alaska for the year 2019. Each image comprises nine spectral bands that cover the visual spectrum (RGB), near-infrared (NIR), and shortwave infrared (SWIR) regions, all harmonized to a 10-meter spatial resolution. 
Although many studies use satellite-derived indices such as NDVI or EVI~\cite{pastick2015distribution}, we do not find consistent performance gains during evaluation and thus solely use raw Sentinel-2 spectral bands in our models. We use the imagery collected during mid-summer, a period that corresponds with peak vegetation growth and provides potentially clearer delineation of environmental boundaries than other seasons. This temporal window also aligns with the time frame of most NSP observations from field surveys. 

\subsection{Topography}
Topographic data refers to rasters that describe surface features on the Earth, including elevation, slope, aspect, and other terrain characteristics, which represent the overall hydrology and geomorphology of a region. These features capture key environmental factors, such as water flow, soil temperature, and erosion, that are essential for predicting NSP presence and broader soil properties~\cite{pastick2015distribution, hengl2017soilgrids250m}. In this paper, we utilize a digital elevation model (DEM), a raster dataset in which each pixel represents the elevation of a location, and compute a variety of terrain derivatives from the base DEM using Whitebox GAT~\cite{lindsay2016whitebox}. All the covariates are represented as rasters at a 10-meter resolution.

The specific topographic layers are as follows. \textbf{Elevation}: A DEM layer that serves as the input for all terrain-related computations. \textbf{Aspect}: The compass direction that a slope faces, derived from elevation gradients. Aspect mainly influences solar radiation exposure and regional climates. \textbf{Maximum Curvature}: Curvature of a location calculated over a 4-meter scale, which reflects the sharpest rate of elevation change. \textbf{Slope}: Slope is also computed at a 4-meter scale, representing the steepness or incline of the terrain. \textbf{Stream Power Index (SPI)}: A hydrological index that quantifies the erosive power of flowing water as a function of slope and upstream contributing area. SPI is useful for identifying areas prone to erosion or channel formation. \textbf{Topographic Position Index (TPI)}: TPI is calculated by subtracting the elevation of each location from the mean elevation of its neighbors within 4 meters. TPI helps classify landscape positions such as ridges or valleys. \textbf{Saga Wetness Index (SWI)}: A variant of the Topographic Wetness Index that indicates the potential for water accumulation by incorporating slope and upstream water accumulation at a 10-meter scale. SWI can be used to infer potential soil moisture and surface saturation. 

% whitebox tools: https://www.whiteboxgeo.com/whitebox-workflows-for-python/

The selection of these covariates is decided based on their relevance to NSP and soil taxonomy prediction tasks. For example, raw elevation is included because of its strong influence on local temperatures and high correlation with NSP distribution~\cite{pastick2015distribution}. Additional DEM-derived variables are chosen for their effect on drainage, solar radiation, and soil moisture \cite{chaplot2000improving, hengl2017soilgrids250m}. While a wide range of topographic derivatives are available, we intentionally restrict our selection to seven DEM-based covariates. The choice strikes a balance between model simplicity and predictive strength, as deriving a large number of features can be computationally expensive, and many topographic variables are highly correlated, which may introduce redundancy or noise into the models.

\subsection{Climate}
Climate plays a fundamental role in shaping both the distribution of permafrost and general soil properties. Covariates such as temperature and precipitation directly influence NSP persistence and soil development processes\cite{pastick2015distribution, thaler2023high}. In this work, we use 30-year normals for three variables, computed by the PRISM Climate Group for the period 1981-2010~\cite{daly20181981} and made available through AKSDB. This time window is aligned with the temporal span of most field observations in MSP dataset. Each raster pixel in the 30-year normals represents the mean value of a climate variable over 30 years. The native resolution of the climate data is 800 meters/pixel, and we interpolate them to 10 meters/pixel to align with other modalities.

The specific climate covariates are as follows. \textbf{Mean Annual Precipitation}: This is the average total yearly precipitation, including rain and snow, expressed in millimeters. This variable indicates soil moisture levels and freeze-thaw potential. \textbf{Summer Warmth Index (SWI)}: This is the sum of mean temperature from May through September, indicating the extent of seasonal thaw that a geographic region may undergo. \textbf{Minimum January Temperature}: This is the average daily minimum temperature for January, which captures the annual coldest conditions over the year at which permafrost may freeze.  

These climate variables are selected based on their documented relevance to NSP and soil taxonomy prediction tasks in the literature. Precipitation, temperature, and thaw indicators like SWI have been identified as important drivers of permafrost dynamics, particularly at a large scale (\textasciitilde1,000 meter resolution) \cite{nelson2003climate}. While additional climate variables are available, many of them are highly correlated with one another. To reduce the potential redundancy and computation needs, we select these three variables for the soil mapping tasks.

% Incorporating both climate variables, which are coarse-scale drivers of permafrost and DEM, which are fine-scale drivers (~10-m), may allow the models to produce maps that can also reflect both coarse and fine scale permafrost patterns. 
\section{\modelname~for Soil Mapping}
\begin{figure}[h]
    \centering
    \includegraphics[width=\linewidth]{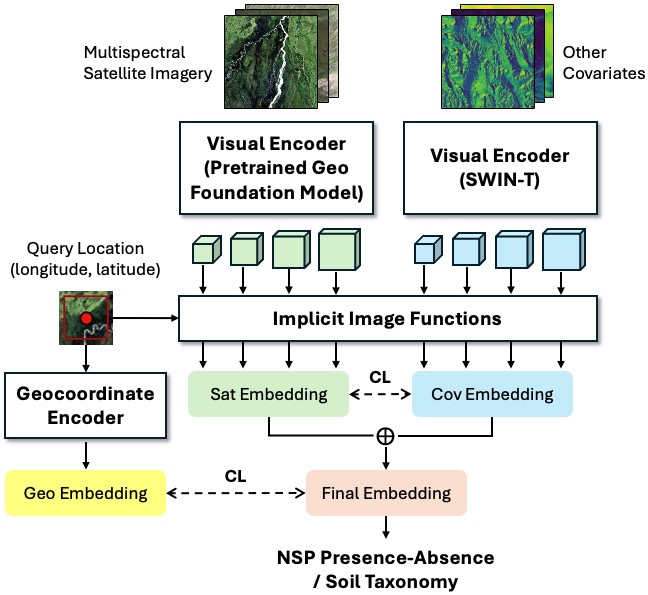}
    \caption{The model architecture of \modelname. The model contains two vision encoders based on SWIN Transformers to handle input modalities and uses implicit image functions to generate continuous spatial predictions for a given geocoordinate. During training, the model employs contrastive learning (CL) objectives to align both multimodal inputs and visual-geospatial representations (indicated by dash arrows). The final embeddings are used to perform downstream tasks, i.e., predicting NSP presence-absence and soil taxonomy.}
    \label{fig:model_architecture}
\end{figure}

This section introduces \modelname, a computer vision model for fine-scale soil mapping. Figure~\ref{fig:model_architecture} shows \modelname, which takes as input cropped multi-channel tiles representing a local region of environmental covariates, along with target location coordinates within it. \modelname~outputs predictions of NSP presence-absence probability or soil taxonomy for each target location. The point-based field observations serve as ground truth labels, guiding the model to learn spatial patterns and feature interactions for prediction.

\subsection{Encode Raster Data}
The input covariates are represented as a cropped image of size $256 \times 256 \times C$, where $C$ denotes the number of input channels (features)\footnote{The image size covers approximately 2.5 km $\times$ 2.5km region.}. We adopt the SWIN Transformer (SWIN-T)~\cite{liu2021swin} to encode the input into multi-scale feature embeddings. Specifically, SWIN-T computes self-attention using non-overlapping local windows and shifted windows, allowing for both within-window and cross-window interactions. As the network progresses, SWIN-T employs a patch merging operation to reduce spatial resolution and increase channel dimension, thereby producing a multi-scale representation. This architecture produces four stages of feature maps with shapes of $64\times64\times128$, $32\times32\times256$, $16\times16\times512$, and $8\times8\times1024$, respectively. These hierarchical embeddings capture rich local and global contextual information, which is important for modeling soil properties governed by both fine-grained patterns and broader environmental gradients. Also, the SWIN-T-based model is selected because of its strong empirical performance, consistently outperforming Vision Transformer-based pretraining models including SatMAE~\cite{cong2022satmae} and Prithvi~\cite{jakubik2023foundation} in soil prediction tasks.

We divide the input modalities into two groups and process them using separate encoders. The first group consists of satellite imagery, for which we adopt SATLASNET~\cite{bastani2023satlaspretrain}, a pretrained SWIN-T-based geospatial foundation model that provides a robust initialization for satellite image representations. The second group includes DEM derivatives and climate variables, which typically lack pretrained models. \modelname~concatenates these channels as one modality and encodes them using a second encoder trained from scratch.

\subsection{Learn Local Implicit Image Function}
After generating image features, a common method for fine-scale prediction is to estimate values for each pixel on a predefined output grid. This method typically requires mapping sparse point measurements to their corresponding grid cells. However, such mapping can be problematic when multiple field observations fall within the same cell and have mixed soil types or permafrost conditions, especially in heterogeneous landscapes like Alaska. To directly handle sparse field observations and avoid predefining a fixed output resolution, an innovation in \modelname~is that it adopts an implicit image function approach~\cite{chen2021learning} used in image super-resolution. This approach enables the model to learn representations (vectors) at arbitrary spatial coordinates through a continuous function over the image feature space.

\begin{figure}[h]
    \centering
    \includegraphics[width=0.6\linewidth]{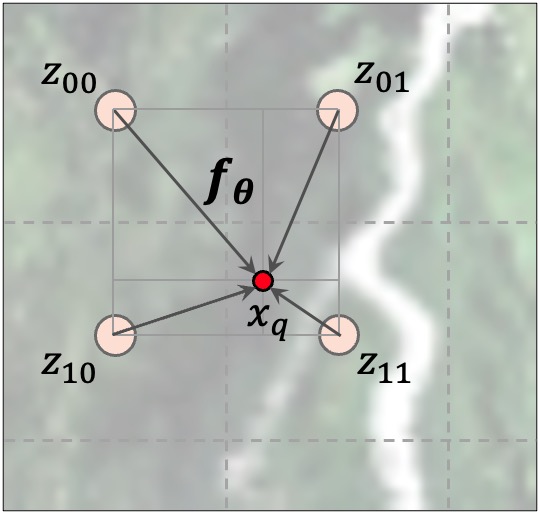}
    \caption{The implicit image function $f_\theta$ learns the feature embedding at an arbitrary query location ({\textcolor{red}{\large$\bullet$}}) from its closest discrete image features ({\textcolor[rgb]{0.98,0.87,0.82}{\large$\bullet$}}). }
    \vspace{-0.1in}
    \label{fig:liif}
\end{figure}

Figure~\ref{fig:liif} illustrates the learning process. Given an $i$-level feature map $Z^{(i)}\in\mathbb{R}^{H\times W\times D}$ from the encoder, and a query location~$x_q$ at within the image, \modelname~aims to learn a continuous feature space~$G^{(i)}$ to represent the context of locations. The embedding at the coordinate~$x_q$ is defined as,
\begin{equation}
G^{(i)}(x_q) = \sum_{t \in \{00, 01, 10, 11\}} \frac{S_t}{S} \cdot f_\theta(z_t, x_q - v_t)
\label{eq:implicit_function}
\end{equation}

Here $x_q$ is a $2D$ coordinate in the continuous image domain. The variable $z_t \in Z^{(i)}, t \in \{00, 01, 10, 11\}$, is the nearest image feature at top-left, top-right, bottom-left, bottom-right to the query location, each associated with a center coordinate $v_t$. The function $f_\theta$, implemented as a multilayer perceptron (MLP), maps discrete feature $z_t$ into a continuous representation by incorporating the relative offset $x_q - v_t$. The four embeddings are then combined using weights proportional to the area $S_t$ of the rectangle between $x_q$ and $v_t$. The weights are normalized by $S=\sum_t S_t$. This weighting scheme ensures that feature locations closer to the query location contribute more to the final embedding.

In this way, \modelname~learns feature representations at arbitrary spatial locations without requiring a predefined output grid or mapping from point observations to grid cells to support fine-scale soil mapping. For each query location, \modelname~computes embeddings from each feature map of the encoder with separate $f_\theta$, aligning the feature dimensions of four scales to the same dimension (D=1,024). These embeddings, extracted from four spatial scales, are then aggregated to form the representation for the query location.
 
\subsection{Learn from Multiple Modalities}
This section introduces the process of learning an embedding to contextualize a query location $x_q$ from multimodal data and incorporate geospatial information.

\subsubsection{Align multiple modalities}
% We divide the input modalities into two groups and process them using separate encoders. The first group consists of satellite imagery, for which we adopt SATLASNET~\cite{bastani2023satlaspretrain}, a pretrained SWIN-T-based geospatial foundation model that provides a robust initialization for satellite image representations. The second group includes DEM derivatives and climate variables, which typically lack pretrained models. \modelname~concatenates these channels as one modality and encodes them using a second SWIN-T encoder trained from scratch.

For a given query location $x_q$, both encoders interact with a set of implicit image functions to produce localized feature embeddings, resulting in two embeddings: $G_{sat}(x_q)$ from the satellite encoder and $G_{cov}(x_q)$ from the other covariate encoder.
To leverage the knowledge from the pretrained model and guide the learning process of the untrained encoder, \modelname utilizes contrastive learning to encourage the alignment between $G_{sat}$ and $G_{cov}$. Specifically, \modelname adopts InfoNCE loss~\cite{oord2018representation} by treating the embeddings for the same location across the two modalities as a positive pair, while embeddings from other locations within the same batch are negative pairs. This multimodal alignment loss enables cross-modal guidance and facilitates the learning of shared information between modalities. The final embedding for each location is obtained by summing the aligned embeddings from the two modalities.

\subsubsection{Enable geo awareness}
Soil properties are strongly associated with geospatial locations, making it essential for learned feature embeddings to be geo-aware. While one could directly concatenate raw coordinates with other input features, this approach might lead models to overly rely on location signals or even produce grid-like prediction patterns, especially when labels exhibit high spatial autocorrelation. Inspired by \cite{mai2023csp}, \modelname uses a coordinate encoder to transform each query location into a geo embedding and utilizes contrastive learning to align this embedding with the corresponding image-derived representation. Similar to the multimodal alignment, we treat embeddings from the same location as positive pairs and those from different locations within the batch as negatives. This alignment encodes geospatial relationships implicitly within the final representation without requiring hard-coded spatial information.

To learn geo embedding, we first define the positional encoding (PE) by applying a Fourier-based transformation to the 2D coordinate ($\lambda, \phi$), representing the longitude and latitude of the query location in EPSG:3338 (Alaska Albers)~\cite{mai2020multi}, computed as:
\[
\mathrm{PE(\lambda, \phi)} = \big[\lambda, \phi, \; \sin(2^k \pi \lambda), \cos(2^k \pi \lambda), \; \sin(2^k \pi \phi), \cos(2^k \pi \phi) \big]_{k=0}^{L-1}
\]
where $L$ is the number of frequencies. We then use three fully connected layers to encode the positional encoding into the target dimension ($D$=1,024). 

% To learn the final embedding for $x_q$, we apply an attention mechanism that uses the geo-coordinate of $x_q$ as query and the multimodal, multi-scale feature embeddings ($G_{sat}$ and $G_{cov}$) as keys and values. In this way, \modelname~learns a spatial adaptive combination of features to account for the spatial heterogeneity across Alaska, where different locations may require different local surface and terrain information. 

\subsection{Training and Inferencing}
During training, we first pretrain \modelname~with only contrastive learning objectives, i.e., alignment between modalities and between visual \& geo embeddings. We randomly crop tiles across Alaska and uniformly sample a fixed number of locations within each tile for pretraining. Then, we fine-tune \modelname~using ground truth labels to optimize the target prediction task. Because the labels are provided as discrete point observations with geographic coordinates (longitude and latitude), our model is well-suited to this input format, enabling flexible spatial querying. We use binary cross-entropy loss for the NSP presence-absence task and cross-entropy loss for the soil taxonomy classification.

During inference, we generate predictions at a user-defined grid resolution (e.g., 10 meters) by defining the query locations as the centers of grid cells. To mitigate boundary artifacts and ensure smooth transitions across adjacent tiles, we use overlapping cropping to extract multi-channel input tiles from the region of interest. Each tile is processed independently by \modelname~to generate predictions. To merge overlapping predictions, we apply a Gaussian weighting scheme, where predictions near the center of each tile are given higher weights. At the same time, those closer to the edge are down-weighted according to a Gaussian kernel. The aggregated results become the final predicted soil map.

\section{Experiments and Results}

\subsection{Soil Prediction Tasks}
\subsubsection{Task 1: Near-Surface Permafrost Presence-Absence.} Task~1 focuses on binary classification, aiming to determine whether NSP is present at a given set of locations. The ultimate objective is to generate a probability map of NSP over the target region. 
In this context, a good predicted map should effectively highlight permafrost presence areas to guide further investigation and planning.

\subsubsection{Task 2: Soil Taxonomy.}
Task 2 aims to predict a multi-class classification for soil taxonomy at a given set of locations. The classification includes seven major soil orders commonly found in Alaska: Andisols, Entisols, Gelisols, Histosols, Inceptisols, Mollisols, and Spodosols. Similarly, the goal is to generate class-specific probability maps across the target region. A good predicted map should accurately capture the spatial heterogeneity of soil types in Alaska.

\subsection{Experiment Settings}
\subsubsection{Training Settings} We divide the field observations into five folds for cross-validation. We use three types of data splits, reflecting different practical scenarios: (1) \textit{Random}: We randomly split data points regardless of locations to assess model performance under ideal sampling conditions~\cite{pastick2015distribution}. (2) \textit{SH-1km}: We apply spatial clustering using DBSCAN~\cite{ester1996density}, grouping observations that fall within a 1 km buffer. We then randomly split these clusters into five folds. This setup simulates the scenario where, for a target location, no neighboring locations have been observed. (3) \textit{SH-10km}: We increase the clustering buffer to 10 km, which creates a larger spatial gap between training and testing regions. This setting evaluates model performance in highly isolated and under-sampled regions.\footnote{Here, ``SH'' stands for Spatial Holdout.}

\subsubsection{Evaluation Metrics} For the binary classification task (Task 1), we evaluate model performance using precision and recall for class 0 (absence) and class 1 (presence). We also compute the overall accuracy, defined as the percentage of correctly predicted labels compared to the ground truth. For the multi-class classification task (Task 2), we evaluate the weighted precision, recall, and F1 scores based on the number of true instances for each class. For each metric, we report the average across five folds.

\subsubsection{Model Settings}
We use a Random Forest model with 253 trees, a minimum sample split of 5, and a maximum tree depth of 16. We used a 100-iteration random search to determine the best-performing hyperparameters set. We experimented with various buffering strategies to aggregate the covariates into a feature vector for Random Forest (RF). For every covariate layer, at a sampled point, we extended a buffer of some distance $d$ and averaged all covariate values within $d$. We tested buffer distances of 0, 25, 50, and 100 meters. We empirically determine that the 50-meter buffer yields the best performance (the RF settings used throughout this paper). In addition to all covariates described in Section~\ref{sec:Datasets}, we also include latitude and longitude as input features.

For \modelname, we pretrained the model for 50 epochs, followed by finetuning on each fold's labels for 30 epochs. During pretraining, we used a learning rate (LR) of 0.0001 and a batch size of 32. During finetuning, we adopted a cosine learning rate schedule with a warm-up LR of 5e-7 and a maximum LR of 5e-5. We conducted all experiments on a single NVIDIA A100 GPU. 

\subsection{Results and Discussion}
Our goal is to compare \modelname with the widely used Random Forest approach by soil scientists to gain a thorough understanding of their respective strengths and weaknesses.

\subsubsection{Quantitative Results}
Table~\ref{tab:permafrost} presents the quantitative comparison between RF and \modelname~for Task 1. \modelname~outperforms RF in \textit{Random} data split. Futher, we observe that, in the ground truth, only 3,408 out of 15,013 locations report permafrost presence, showing a strong class imbalance towards the absence class. Despite this, although \modelname~has lower overall accuracy than RF in \textit{SH-1km} and \textit{SH-10km} splits, \modelname~consistently achieves higher recall for permafrost presence (``R1'') compared to RF, indicating \modelname~produces conservative probability estimates while RF tends to favor the majority class, resulting in higher overall accuracy but lower sensitivity to potential permafrost than \modelname. This conservative behavior of \modelname~ is extremely valuable for domain scientists, particularly in guiding fieldwork and early-warning assessments, where missing areas with possible permafrost may have greater consequences than over-predicting them. We further examine the regions where \modelname~produces false positives. Some areas were historically underlain by permafrost but recently thawed due to the global warming trends. However, surface characteristics in satellite imagery may not yet accurately reflect these subsurface changes, suggesting that remote sensing data could serve as a delayed indicator of the real-time conditions.

\begin{table}[h!]
\centering
\caption{Performance comparison between RF and \modelname~on NSP presence-absence across three data splits.}
\begin{tabular}{lcccccc}
\toprule
\textbf{5 Folds} & \textbf{Method} & \textbf{P0} & \textbf{R0} & \textbf{P1} & \textbf{R1} & \textbf{Acc} \\
\midrule
\multirow{2}{*}{\centering Random} & RF    & 95.41 & 98.08 & 92.82 & 83.91 & 94.87 \\
                                   & \modelname  & {97.38} & {98.26} & {93.90} & {90.98} & {96.61} \\
\midrule
\multirow{2}{*}{\centering SH-1km} & RF    & 93.46 & {96.65} & {87.12} & 77.19 & {92.20} \\
                                    & \modelname  & {95.33} & 92.17 & 76.17 & {84.87} & 90.48 \\
\midrule
\multirow{2}{*}{\centering SH-10km} & RF    & 88.29 & {91.16} & {69.29} & 63.07 & {83.44} \\
                                    & \modelname  & {91.27} & 82.68 & 58.62 & {75.93} & 81.06 \\
\bottomrule
\end{tabular}
\label{tab:permafrost}
\end{table}

\begin{figure}[h]
    \centering
    \includegraphics[width=\linewidth]{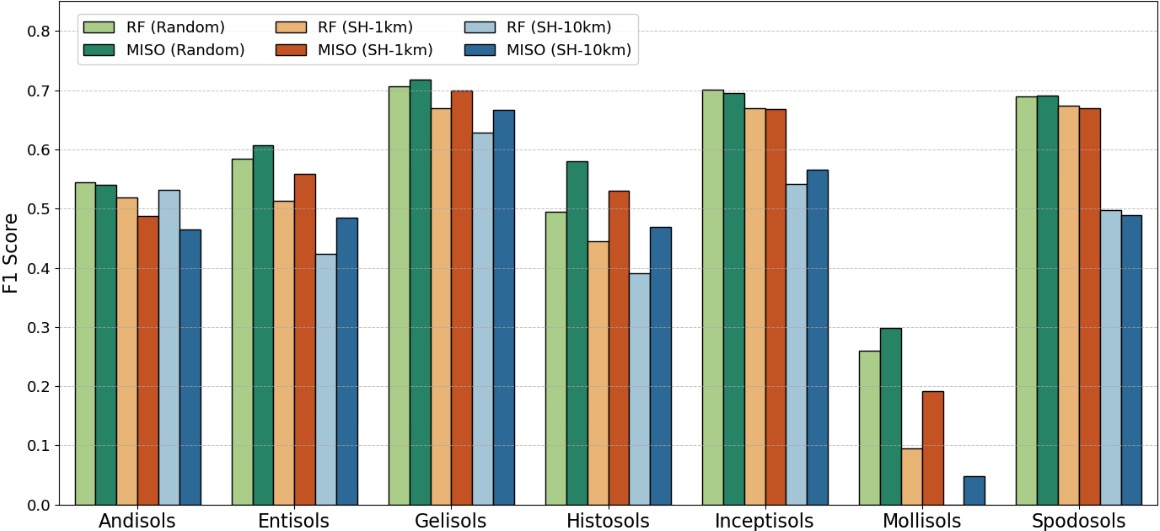}
    \caption{Comparison between RF and \modelname~across soil taxonomy and data splits}
    \vspace{-0.1in}
    \label{fig:tax}
\end{figure}

Table~\ref{tab:tax} presents the quantitative results for Task 2.  Overall, \modelname~consistently outperforms RF across three data splits. Similar to Task 1, \modelname generally achieves higher recall than RF, and it also yields higher precision. One potential reason is that Task 2 includes 32,132 field observations, more than double the size of Task 1 and covering wider locations.
Figure~\ref{fig:tax} shows class-level F1 scores for soil taxonomy prediction. We observe that \modelname~achieves better performance than RF in most soil types, including  Entisols, Gelisols, Histosols, and Mollisols. In particular, the improvements on Entisols, Gelisols, and Histosols demonstrate the strength of \modelname's capability in extracting important signals from imagery data. These soil types are closely related to observable surface features captured by satellite imagery. For example, Histosols---organic-rich, water-saturated soils---are commonly found in dark-surface wetlands, floodplains, or areas near rivers and lakes with high moisture signals. 
Furthermore, Mollisols, which account for only 1.5\% of the dataset, unsurprisingly have the lowest performance among all soil types due to severe class imbalance. Nonetheless, \modelname’s advantage here indicates a lower tendency to collapse predictions toward majority classes compared to RF. 
For Inceptisols and Spodosols, \modelname~and RF perform comparably well since these soil types have consistent environmental signals. 
% Inceptisols, for example, tend to occur in moderately developed mountain soils and are common across mid- to high elevations in Alaska. Spodosols are common in cool, humid climates with evergreen forest cover. 
Lastly, for Andisols, RF outperforms \modelname. Andisols are volcanic soils often found in high-elevation regions.
One potential reason is that \modelname~might learn misleading surface patterns in volcanic areas that do not correlate with Andisols' presence.

\begin{table}[h!]
\centering
\caption{Performance comparison between RF and \modelname~on soil taxonomy classification across three data splits.}
\begin{tabular}{lccccc}
\toprule
\textbf{5 Folds} & \textbf{Method} & \textbf{Precision} & \textbf{Recall} & \textbf{F1} \\
\midrule
\multirow{2}{*}{\centering Random} & RF    & 65.47 & 65.53 & 64.67 \\
                                   & \modelname  & 66.16 & 66.31 & 65.89 \\
\midrule
\multirow{2}{*}{\centering SH-1km} & RF    & 61.83 & 62.14 & 60.83 \\
                                   & \modelname  & 63.01 & 63.44 & 62.77 \\
\midrule
\multirow{2}{*}{\centering SH-10km} & RF    & 53.29 & 52.67 & 54.88 \\
                                    & \modelname  & 55.86 & 55.61 & 55.11 \\
\bottomrule
\end{tabular}
\label{tab:tax}
\end{table}

\subsubsection{Predicted Probability Distribution}

% We further examine the predicted probability distribution. Figure~\ref{fig:task1_prob_distribution} shows the permafrost probability distribution combining all testing folds (Task 1). We compute the percentage of the prediction in every probability bins. We observe that both methods have a bi-modal trend where the prediction are concentrated in 0-0.1 and 0.9-1.0 bins. The predictions in 0-0.1 bin is way more than other bins, which is consist with the fact that there are more non-permafrost observations than permafrost in the original field observation dataset. Also, \modelname has a more sharp prediction results than RF where there are rare predictions located in the middle probability value. The probabilities sitting in the middle might be useful to decide the new sampling locations as thesa locations are quite ambiguity to the models. While \modelname has fewer such points than RF and classfiy them into 0 or 1 with high confidence. 

% Figure~\ref{fig:task2_G_prob_distribution} shows the probability distribution for one of the soil types, Gelisols which is the dominant soil in Alaska. We observe that \modelname still behaves a bi-modal distribution in the predictions. In comparison, RF has a decreasing trend in the probability of Gelisols, where the model tends to predict small probability while \modelname~produces more confident, bimodal predictions with probabilities closer to 0 or 1. 

\begin{figure}[h]
    \centering
    \begin{subfigure}[t]{0.9\linewidth}
        \centering
        \includegraphics[width=\linewidth]{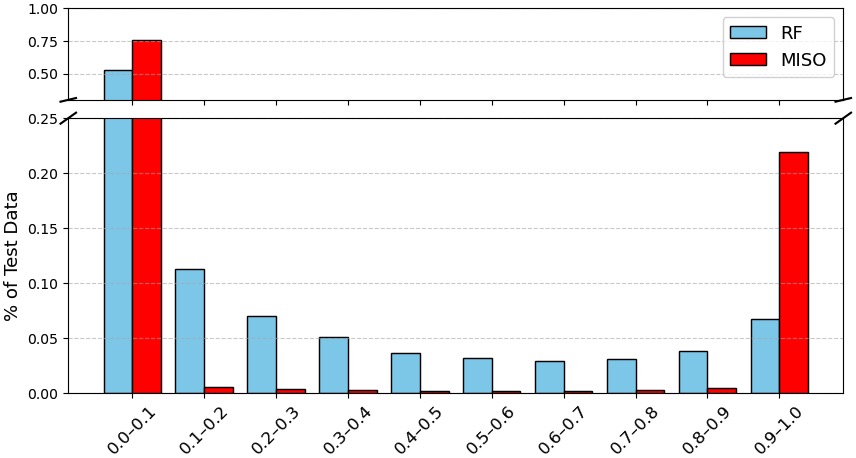}
        \caption{Task 1: NSP Probability Distribution}
        \vspace{0.1in}
        \label{fig:task1_prob_distribution}
    \end{subfigure}
    \begin{subfigure}[t]{0.9\linewidth}
        \centering
        \includegraphics[width=\linewidth]{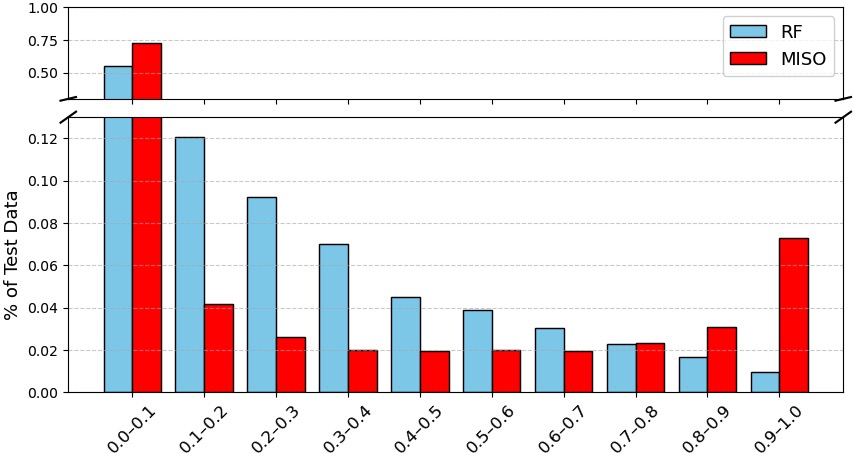}
        \caption{Task 2: Gelisols Probability Distribution}
        \label{fig:task2_G_prob_distribution}
    \end{subfigure}
    \caption{Predicted probability distribution of RF and \modelname~in two tasks}
    \vspace{-0.1in}
    \label{fig:prob_distribution_combined}
\end{figure}

From here, we focus on the results from \textit{SH-1km} data split as we find this to be a practical scenario for many end users. We examine the predicted probability distributions by merging the predictions from all test folds. We group the predictions into bins based on probability values and compute the normalized count within each bin. Figure~\ref{fig:task1_prob_distribution} illustrates the distribution of NSP probabilities. The overall trend generally reflects the class distribution in the field observations, where non-NSP samples are prevalent. Both models exhibit a bimodal distribution, with probabilities heavily concentrated in the 0-0.1 and 0.9–1.0 ranges. Specifically, \modelname~produces a sharper distribution than RF with fewer probabilities in the intermediate bins (0.2–-0.8). This phenomenon may limit the model's utility for identifying geographic regions where the output is uncertain, but it generally improves overall classification performance on potential NSP.
%highlights a trade-off: While \modelname’s confident predictions may improve classification on potential NSP, the mid-range probabilities may be important cues for identifying regions with high uncertainty for further investigation.

Figure~\ref{fig:task2_G_prob_distribution} shows the probability distribution for one of the soil classes, Gelisols, predicted in Task 2. \modelname~again displays a bimodal distribution, reinforcing its tendency to produce decisive predictions. In contrast, RF exhibits a monotonically decreasing distribution, suggesting RF might have difficulty differentiating Gelisols from other soil types in many locations. 

\subsubsection{Spatial Analysis}
To further investigate the spatial behavior of model predictions, we perform a detailed analysis using task-specific geographic partitions for the two tasks.  
\vspace{0.08in}

\noindent \textbf{Permafrost Zones and Permafrost Presence-Absence.} 
Permafrost zones are delineated by domain experts to categorize permafrost patterns in Alaska. We examine the prediction results of RF and \modelname~against an existing permafrost zone map~\cite{jorgenson2008permafrost} of Alaska that splits the state into continuous (>90\% of land has permafrost), discontinuous (50-90\%), sporadic (10-50\%), isolated (0-10\%), unfrozen, glacier, and waterbody zones (seeFigure~\ref{fig:permafrost}). In our assessment, we do not use glacier and waterbody zones due to their low observation counts and also exclude unfrozen because almost all points within unfrozen do not contain permafrost. Our aim is to assess how well the predictions from machine learning models align with established knowledge about permafrost distribution and how their results vary within each zone.

%RF and \modelname~have similar overall accuracy trends across permafrost zones, as shown in Table~\ref{tab:pzone}. Both models unsurprisingly perform worst in the discontinuous zone due to its complex environment --- a highly mixed presence-absence region, where permafrost is prevalent but not ubiquitous and exhibits high spatial heterogeneity. One detail to note is that RF outperforms \modelname~by the largest margin in the continuous zone with the \textit{SH-1km} split. This may be because, despite lower spatial autocorrelation for NSP in the \textit{SH-1km} split than \textit{Random}, RF still heavily relies on latitude and often assigns the same label to points at similar latitudes, and all points in the continuous region occur at high latitudes. In contrast, \modelname does not explicitly use longitude/latitude as input covariates, and predicts false positives at the boundary regions of continuous and discontinuous zones.

% the overall trends in each zone are similar between RF and \modelname. Both models have the lowest accuracy in the discontinuous zone, although \modelname struggles more in the continuous zone in comparison to RF. The large difference in the discontinuous zone may be due to an increase in the difficulty of predicting permafrost in this zone, as permafrost is prevalent but not ubiquitous and may exhibit high levels of spatial heterogeneity. 

\begin{figure}[h]
    \centering
    \includegraphics[width=\linewidth]{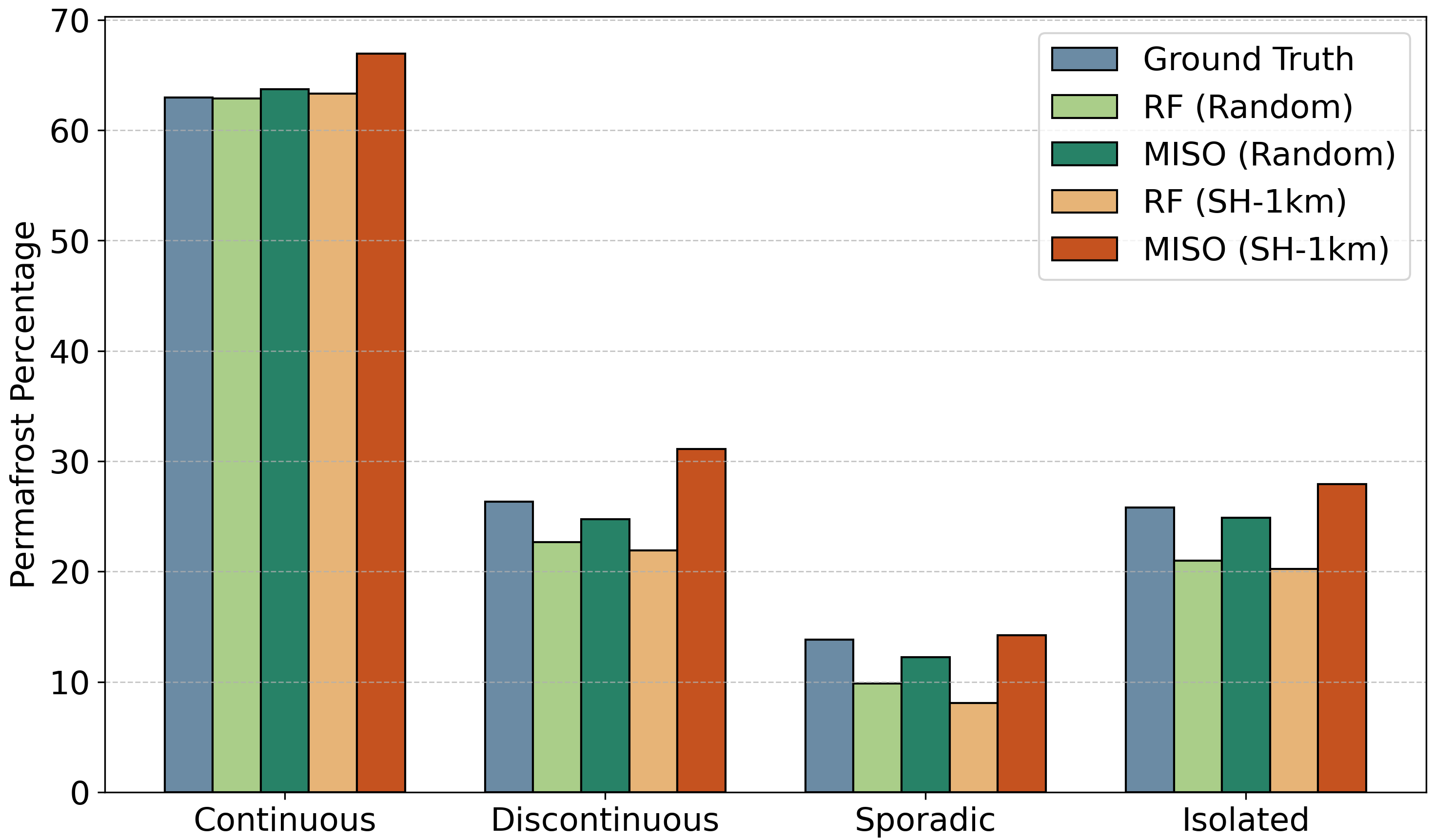}
    \caption{Predicted permafrost presence percentage by Permafrost Zones by RF and \modelname~in both \textit{Random} and \textit{SH-1km} split experiments.}
    \label{fig:zonecomp}
\end{figure}

We focus on the comparison of an important domain criterion -- the predicted NSP presence percentage in each zone (see Figure~\ref{fig:zonecomp}). First of all, it is worth noting that, even in the ground truth of permafrost percentage, there is a significant underestimation that are expected in each zone (i.e., the ground truth percentage in our AKSDB dataset is \textasciitilde60\% in the continuous zone but is expected to be \textasciitilde90\%). This phenomenon has been observed in other studies that map permafrost in Alaska \cite{pastick2015distribution}, which can be attributed to a variety of factors, including sampling errors and spatial biases in the ground truth, as well as localized factors that were not fully accounted for when the permafrost zones were defined. 

RF typically underestimates the amount of NSP in all zones except the continuous zone across both the \textit{Random} and \textit{SH-1km} splits. This is likely related to RF's tendency to overestimate the majority class. While the results in both splits are similar for the RF model, \modelname, conversely, will overestimate permafrost percentage with the \textit{SH-1km} split and slightly underestimate with the \textit{Random} split. This highlights the significance of the training set's spatial distribution on the biases in the results. When using \modelname, if the training set is evenly distributed across the geographic space, end users may find that the result slightly underestimates the amount of permafrost present. At the same time, the opposite may occur if the training set contains spatial clusters of points that are far from the inference set.

Both models exhibit trends that generally reflect pre-existing large-scale permafrost generalizations; for instance, they both show higher predicted permafrost percentages in the continuous and discontinuous zones compared to the isolated and sporadic zones. However, those looking to apply these fine-scale soil products may want to be aware that they do not strictly align with larger-scale permafrost zonal definitions. If end users want to align soil maps with pre-existing domain knowledge, they may find it useful to use the probability maps and set their own thresholds on which probabilities should indicate the presence of permafrost.

% \begin{table}[h!]
% \centering
% \caption{Accuracy by Permafrost Zone. C is continuous, D is discontinuous, S is sporadic, and I is isolated.}
% \begin{tabular}{lccccc}
% \toprule
% \textbf{5 Folds} & \textbf{Method} & \textbf{C} & \textbf{D} & \textbf{S} & \textbf{I} \\
% \midrule
% \multirow{2}{*}{\centering Random} & RF   & 94.37 & 91.25 & 95.43 & 93.90 \\
%                                    & \modelname & 95.31 & 94.30 & 97.20 & 96.98 \\
% \midrule
% \multirow{2}{*}{\centering SH-1km} & RF   & 90.15 & 87.48 & 93.67 & 90.30 \\
%                                    & \modelname & 85.70 & 84.72 & 91.48 & 90.30 \\
% \bottomrule
% \end{tabular}
% \label{tab:pzone}
% \end{table}

\noindent \textbf{Major Land Resource Area (MLRA) and Soil Taxonomy.} MLRAs are geographically distinct regions that have been delineated with attributes such as climate, soil type, land use, vegetation, geology, or physiography \cite{usda2022mlra}. They provide a framework for dividing a large geographic region into sub-regions with generally similar environmental profiles. Since MLRAs are used by soil scientists and land managers to guide decision-making at regional scales, understanding model performance within each MLRA can reveal where predictions are more or less reliable, and help users interpret results in an ecologically meaningful context. To this end, we examine the overall accuracy by class for the soil taxonomy task in each MLRA.
% This, in turn, can inform soil scientists interested in modeling gelisols and their associations with permafrost on what locations to survey.

\begin{figure}[h]
    \centering
    \includegraphics[width=\linewidth]{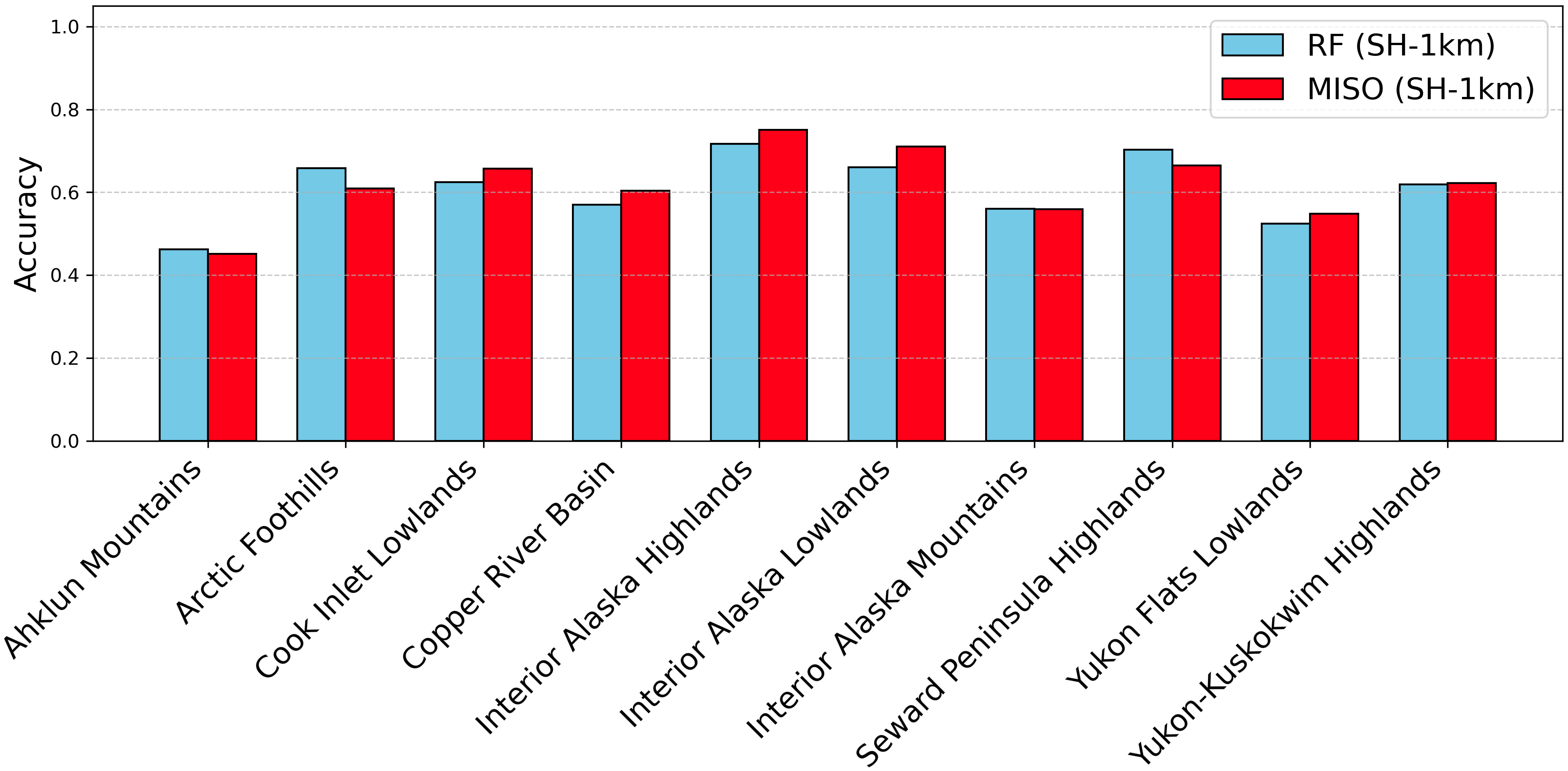}
    \caption{Accuracy comparison between RF and \modelname~in the \textit{SH-1km} split experiment for selected MLRAs.}
    \label{fig:overall_acc_mlra}
\end{figure}

Although \modelname~generally outperforms RF on soil taxonomy classification, \modelname does not uniformly outperform RF in every MLRA, as shown in Figure~\ref{fig:overall_acc_mlra}. We observe that the regions where \modelname has a better overall accuracy than RF have varied characteristics and are spatially disparate. For example, Yukon Kuskowim Highlands in western Alaska, Northern Brooks Range Mountains in northern Alaska, and Cook Inlet Lowlands (CIL) in southern Alaska are all regions where \modelname has more than 4\% improvement in accuracy over RF, but all three are on opposite corners of Alaska, highlighting \modelname's ability to predict soil taxa across different environments and geographic locations. Additionally, these regions all have a diverse number of training points, with areas like the Northern Brooks Range Mountains having only 55 while the Cook Inlet Lowlands contain 4,809, showing that \modelname can outperform RF regardless of training data count. 

In MLRAs where \modelname~generally outperforms RF, there is no single soil class that consistently drives the improvement. Rather, the class with the highest accuracy varies heavily by region. For example, Table~\ref{tab:mlraex} shows that both models achieve high accuracy for Spodosols in CIL but perform poorly in IAH, potentially due to the diverse behaviors of Spodosols in these two regions. Neither model generalizes well for Spodosols across the two geographically distant regions. In contrast, for Gelisols, both models show similar performance in IAH, but \modelname~performs at 50\% accuracy while RF achieves none in CIL, highlighting \modelname's superior capability in handling spatial heterogeneity in soil types.

\begin{table}[h!]
\centering
\caption{Example of accuracy by soil classes in the Cook Inlet Lowlands (CIL) and Interior Alaska Highlands (IAH) in \textit{SH-1km} experiments. The table header indicates (A, Andisols), (E, Entisols), (G, Gelisols), (H, Histosols), (I, Inceptisols), (S, Spodosols). We do not show Mollisols as the number of points is below five in both regions. }
% \resizebox{\linewidth}{!}{
\begin{tabular}{lccccccc}
\toprule
\textbf{MLRA} & \textbf{Method} & \textbf{A} & \textbf{E} & \textbf{G} & \textbf{H} & \textbf{I} & \textbf{S} \\
\midrule
\multirow{2}{*}{\centering CIL} & RF    & 3.8 & 42.1 & 0.0 & 18.2 & 17.7 & 91.7\\
    & \modelname  & 13.2 & 49.9 & 50.0 & 47.2 & 17.5 & 88.5\\
\midrule
\multirow{2}{*}{\centering IAH} & RF    & 0.0 & 30.1 & 70.8 & 0.0 & 86.9 & 0.0 \\
    & \modelname  & 0.0 & 47.0 & 78.6 & 13.0 & 83.9 & 6.25\\
\bottomrule
\end{tabular}
% }
\label{tab:mlraex}
\end{table}

\begin{figure}[h]
    \centering
    \includegraphics[width=\linewidth]
    {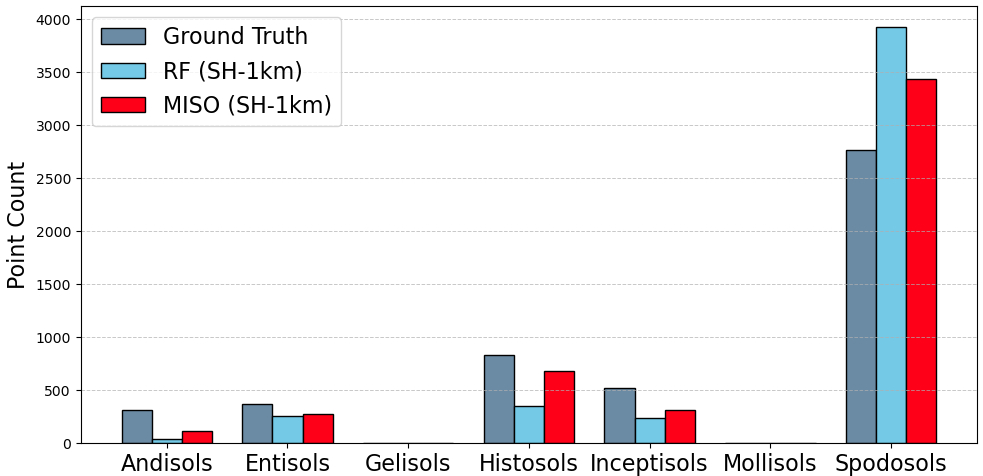}
    \caption{Distribution of the ground truth points, RF predicted point count, and the \modelname~ predicted point count in each soil taxonomy in the Cook Islands Lowlands MLRA.}
    \label{fig:dists_mlra}
\end{figure}

When examining regions with a large number of training points (more than 2,000), we observe that \modelname~tends to have better accuracy than RF and better reflect the distribution of soil taxonomy when in regions where the distribution of classes skews towards one class. In particular, in these regions, RF performs well on the dominant class in that MLRA but poorly on all other classes. RF typically overestimates the number of points that should be in the majority class. In these regions, \modelname~improves the performance on classes that have fewer points in the MLRA and better models the overall soil profile. In Figure
~\ref{fig:dists_mlra} shows the comparison of the ground truth distribution against the distributions of RF and \modelname in the Cook Island Lowlands. The number of predicted Spodosols, the class that is most frequently present in the region, is greatly overestimated in RF, and classes with fewer points, such as Histosols, are all underestimated. In comparison, \modelname~still overestimates Spodosols but to a lesser degree than RF and also has more predicted points in the other classes, demonstrating a distribution of predicted points much closer to the ground truth. This effect is also reflected in Table~\ref{tab:mlraex} where, in both the Cook Inlet Lowlands and Interior Alaska Highlands, RF has better accuracy than \modelname~ in one class (the class with the largest amount of points) and has worse accuracy in all other classes.

As it is a common phenomenon for ecological subunits to have a profile where there is a dominant soil type but the ecoregion still contains heterogeneous soils \cite{shirazi2003quantitative}, end users, particularly those who are interested in only studying a specific MLRA or a localized sub-region of Alaska, applying the taxonomic map may find this analysis useful. \modelname's ability to more accurately model the entire soil profile, rather than only the dominant class, makes it a preferred choice in most soil science analysis scenarios.
For instance, researchers have emphasized the importance of accurately modeling soil heterogeneity in Arctic regions, as it can be essential for understanding soil dynamics in microclimates, which is helpful in predicting permafrost thaw at a fine scale (\textasciitilde50 meters)~\cite{yi2018characterizing}.

\subsubsection{Visualizations} 
\begin{figure}[h]
    \centering
    \begin{subfigure}[t]{\linewidth}
        \centering
        \includegraphics[width=0.8\linewidth]{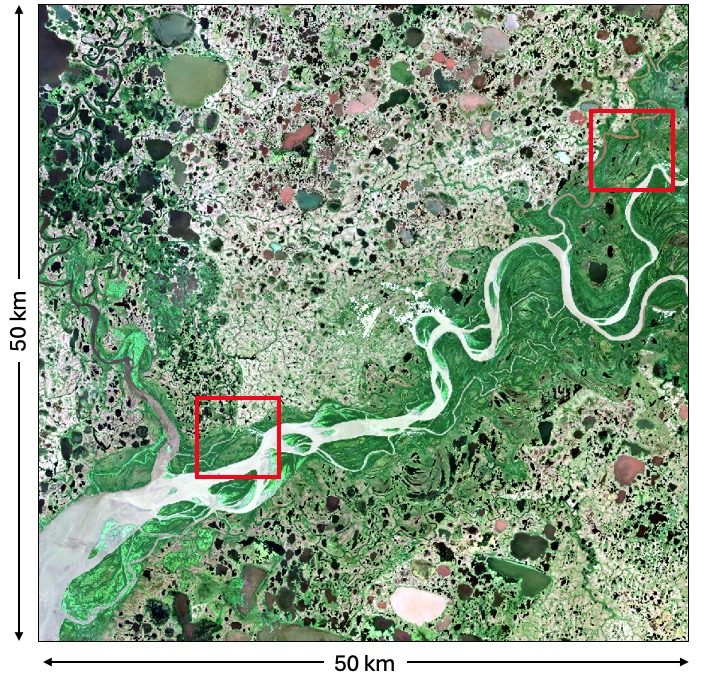}
        \caption{Sentinel-2 RGB Channels enhanced and visualized by QGIS}
        \vspace{0.1in}
        \label{fig:AK050H36V24}
    \end{subfigure}
    \centering
    \begin{subfigure}[t]{\linewidth}
        \centering
        \includegraphics[width=\linewidth]{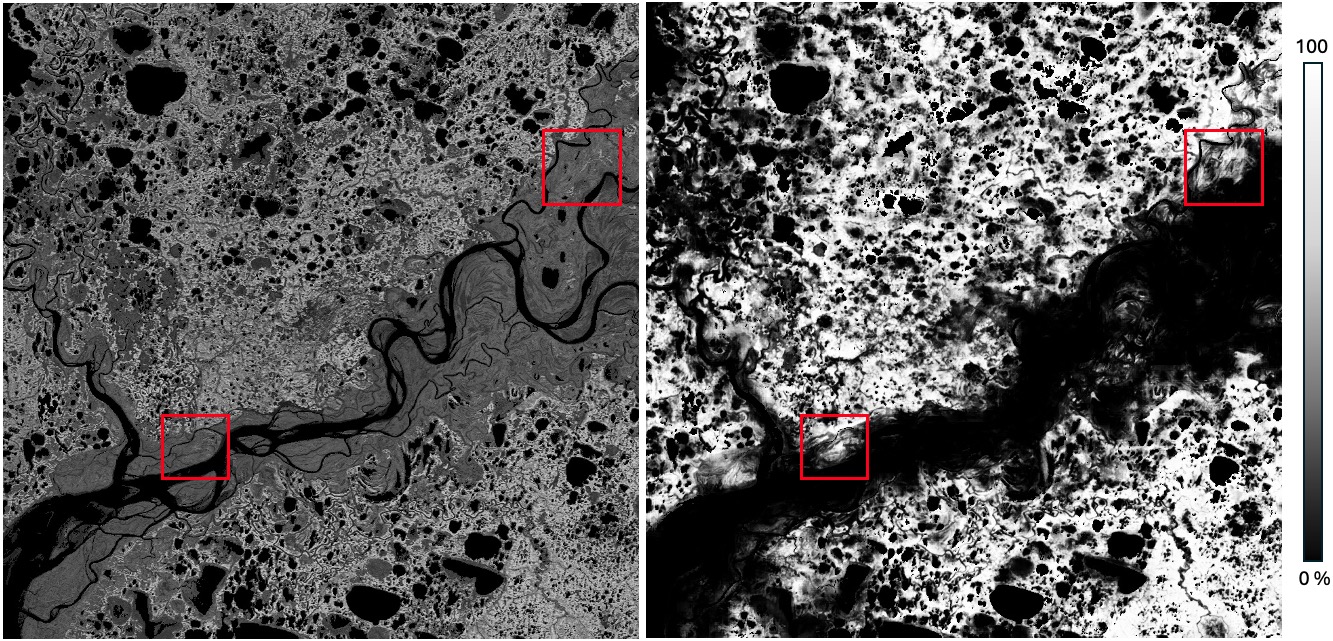}
        \caption{NSP Probability by RF (left) and \modelname~(right)}
        \label{fig:AK050H36V24_PF}
    \end{subfigure}
    \caption{Visual example of fine-scale prediction for NSP}
    \label{fig:soil_fine}
\end{figure}

\begin{figure}[h]
    % \centering
    % \begin{subfigure}[t]{\linewidth}
    %     \centering
    %     \includegraphics[width=\linewidth]{figures/AK050H36V24_PF.png}
    %     \caption{Gelisols Distribution}
    %     \label{fig:AK050H36V24_PF}
    % \end{subfigure}
    % \centering
    % \begin{subfigure}[t]{\linewidth}
    %     \centering
    %     \includegraphics[width=\linewidth]{figures/AK050H36V24_Gelisols.png}
    %     \caption{Gelisols}
    %     \label{fig:AK050H36V24_Gelisols}
    % \end{subfigure}
    \begin{subfigure}[t]{\linewidth}
        \centering
        \includegraphics[width=\linewidth]{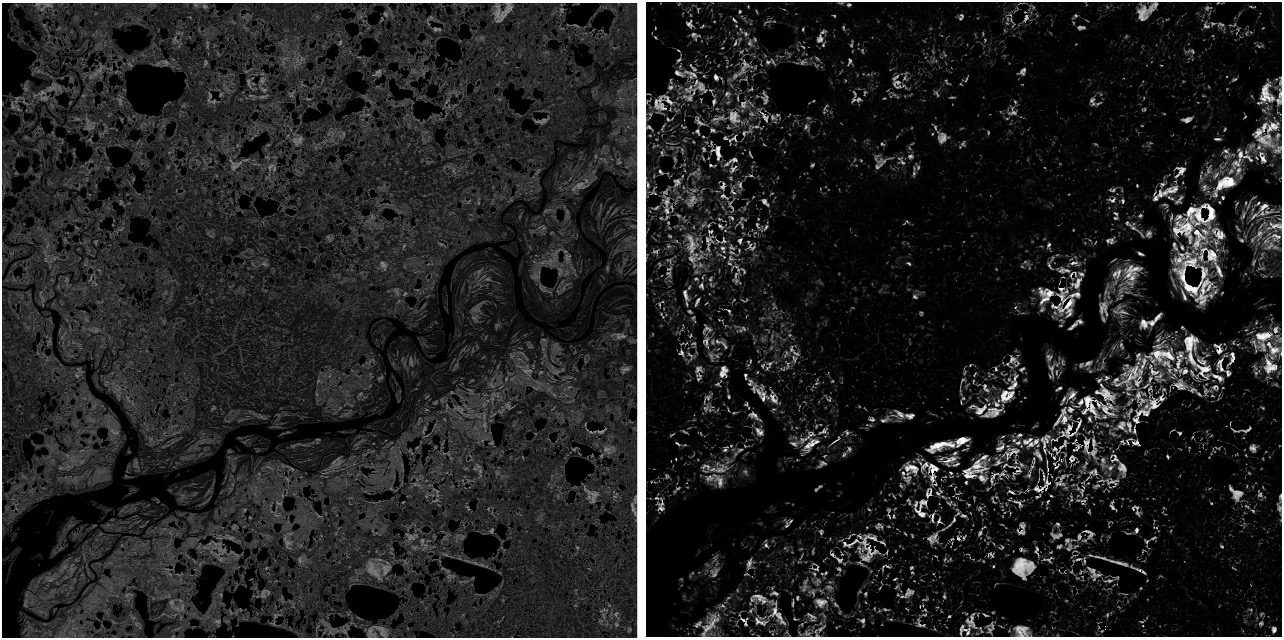}
        \caption{Histosols Probability by RF (left) and \modelname~(right)}
        \vspace{0.1in}
        \label{fig:AK050H36V24_Histisols}
    \end{subfigure}
    \begin{subfigure}[t]{\linewidth}
        \centering
        \includegraphics[width=\linewidth]{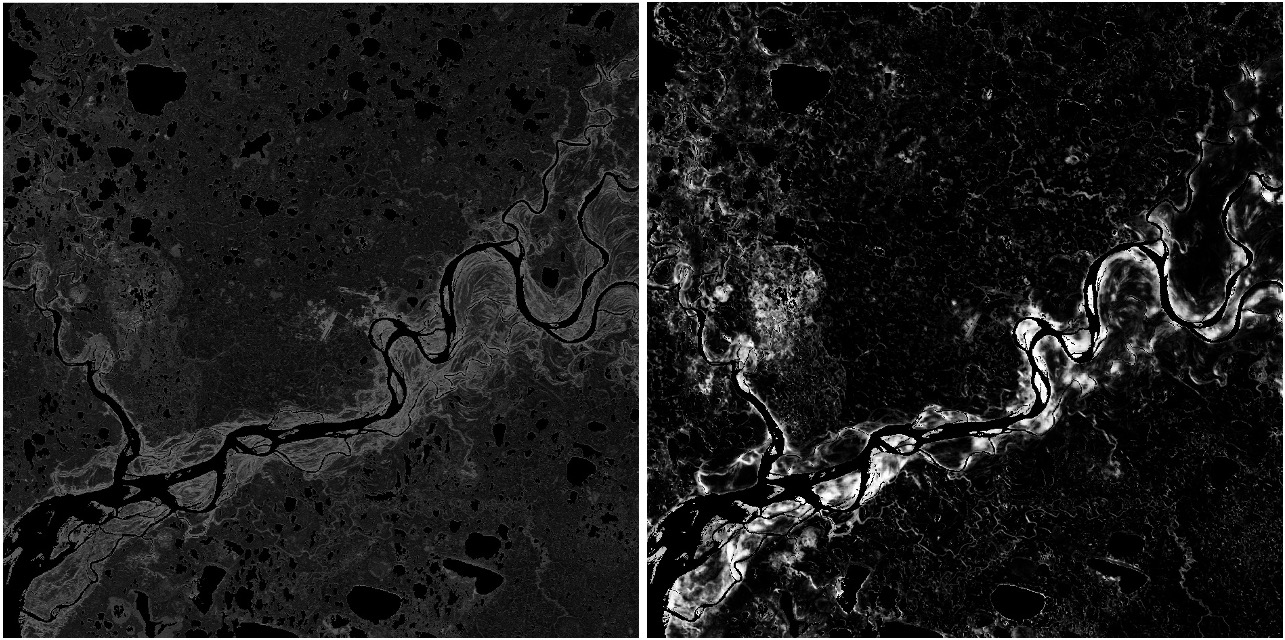}
        \caption{Entisols Probability by RF (left) and \modelname~(right)}
        \label{fig:AK050H36V24_Entisols}
    \end{subfigure}
    \caption{Visual example of fine-scale prediction for soil taxonomy}
    \vspace{-0.1in}
    \label{fig:soil_fine}
\end{figure}

We visually assess the fine-scale predictions for NSP and soil taxonomy in a selected region of interest using RF and \modelname. The target region spans 50 km $\times$ 50 km, which lies in a highly dynamic lowland floodplain in southwest Alaska (Figure~\ref{fig:AK050H36V24}). The landscape features extensive river braiding, dense shallow ponds, and saturated wetlands in the northern and western parts, as well as permafrost-affected areas in the north. Figure~\ref{fig:AK050H36V24_PF} shows NSP predictions where white indicates a higher permafrost probability. Both models capture the general spatial patterns of NSP in the region. When referenced against the satellite imagery, RF predicts permafrost where there is a visual contrast and clear boundary in the satellite imagery (see the red highlighted regions in Figure~\ref{fig:AK050H36V24_PF} for an example), but \modelname~ predicts permafrost beyond this boundary, further down the floodplain (see corresponding red boxes in Figure~\ref{fig:AK050H36V24_PF}). The disagreements between the two models are good candidates for further examination and fieldwork. The visualization shows that \modelname~predicts near-zero probabilities in saturated wetlands along the river, highlighting a sharper boundary between permafrost and non-permafrost areas compared to the RF, which assigns nonzero probabilities in the same regions.
Figure~\ref{fig:AK050H36V24_Histisols} and \ref{fig:AK050H36V24_Entisols} show predictions for two soil classes as an example: Entisols and Histosols. Both models correctly locate these soils in their typical geomorphic settings. Histosols are organic-rich soil and often form in saturated wetlands. \modelname~produces spatially detailed and scattered predictions in the northwestern region, whereas RF shows a broader, lower-confidence response. Entisols, typically found in fluvial environments, are well-aligned along the braided river system in both models. \modelname~again shows sharper transitions and clearer boundaries than RF, potentially due to its stronger spatial contextual features derived from visual cues.

\section{Conclusion, Limitations, and Future Work}
This paper supports fine-scale soil mapping in Alaska with \modelname, a computer vision model specifically adapted for soil property prediction tasks, producing the first ever 10 meter/pixel soil maps of NSP and taxonomy over Alaska and a thorough comparison of \modelname~ against the commonly used RF. While models achieved ~80-90\% accuracy on the NSP prediction task, \modelname produces more conservative permafrost probability estimates. This is especially valuable for identifying areas potentially vulnerable to permafrost thaw and informing the risk assessment process for those who live in such areas or are otherwise affected by thaw. For soil taxonomy, \modelname~demonstrates a strong ability to capture spatially heterogeneous soil types. The soil taxonomy map is valuable in providing fundamental knowledge of soil composition and can be used as a base layer to inform studies on other attributes of the soil profile, such as organic carbon or soil moisture~\cite{billings2021soil}. The detailed analysis of the prediction results, including the effect of geographical areas, the prediction probability distributions, and how the training data influences the overall accuracy, informs users of the produced soil map product and offers a starting point for future improvements in predictions. 

Our work establishes an initial bridge between computer vision and soil science, opening numerous opportunities for future research. One of the most pertinent tasks in soil modeling is forecasting future permafrost presence or thaw under changing climate conditions. While this study does not explicitly account for the time at which points were sampled, future work will incorporate a temporal element to address the forecasting task. 

% Additionally, soil mapping models frequently incorporate numerous different types of covariates, which are multi-sensor and multi-resolution; future work will include further investigation on how to better combine covariate data. 

\bibliographystyle{ACM-Reference-Format}
\bibliography{ref}
\end{document}